\documentclass[11pt,a4paper]{article}
\usepackage[hyperref]{emnlp2018}
\usepackage{times}
\usepackage{latexsym}
\usepackage{amsmath, amsthm, amssymb}
\usepackage{mathtools,xparse}
\usepackage{booktabs}

\DeclareMathOperator*{\argmin}{argmin}
\DeclareMathOperator*{\fro}{fro}
\DeclareMathOperator*{\card}{card}

\usepackage{url}

\aclfinalcopy 

\title{Evaluating Lottery Tickets Under Distributional Shifts}
\author{
    \textbf{Shrey Desai}$^{*,1}$\quad\textbf{Hongyuan Zhan}$^2$\quad\textbf{Ahmed Aly}$^2$\\
    $^1$The University of Texas at Austin \quad $^2$Facebook Assistant\\
    \texttt{shreydesai@utexas.edu}\\
    \texttt{\{hyzhan, ahhegazy\}@fb.com}
    }
\date{}

\begin{document}
\maketitle

\renewcommand{\thefootnote}{\fnsymbol{footnote}}
\footnotetext[1]{Work done during an internship at Facebook.}
\renewcommand{\thefootnote}{\arabic{footnote}}

\begin{abstract}
The Lottery Ticket Hypothesis \cite{frankle2018the} suggests large, over-parameterized neural networks consist of small, sparse subnetworks that can be trained in isolation to reach a similar (or better) test accuracy. However, the initialization and generalizability of the obtained sparse subnetworks have been recently called into question. Our work focuses on evaluating the initialization of sparse subnetworks under distributional shifts. Specifically, we investigate the extent to which a sparse subnetwork obtained in a source domain can be re-trained in isolation in a dissimilar, target domain. In addition, we examine the effects of different initialization strategies at transfer-time. Our experiments show that sparse subnetworks obtained through lottery ticket training do not simply overfit to particular domains, but rather reflect an inductive bias of deep neural networks that can be exploited in multiple domains.
\end{abstract}

\section{Introduction}

Recent research has suggested deep neural networks are dramatically over-parametrized. In natural language processing alone, most state-of-the-art neural networks have computational and memory complexities that scale with the size of the vocabulary. Practitioners have developed numerous methods to reduce the complexity of these models---either before, during, or after training---while retaining existing performance. Some of these methods include quantization \cite{gong2014compressing, hubara2017quantized}, and different flavors of pruning \cite{zhu2017prune,liu2018rethinking,frankle2018the,gale2019state}.

In particular, the Lottery Ticket Hypothesis \cite{frankle2018the} proposes that small, sparse subnetworks are embedded within large, over-parametrized neural networks. When trained in isolation, these subnetworks can achieve commensurate performance using the same initialization as the original model. The lottery ticket training procedure is formalized as an iterative three-stage approach: (1) train an over-parametrized model with initial parameters $\theta_0$; (2) prune the trained model by applying a mask $m \in \{0,1\}^{|\theta|}$ identified by a sparsification algorithm; (3) reinitialize the sparse subnetwork by resetting its non-zero weights to the initial values ($m \odot \theta_0$) and retrain it. These three stages are repeated for multiple rounds. If the final subnetwork achieves similar (or better) test performance in comparison to the original network, a winning \textit{lottery ticket} has been identified. 

Evidence of the existence of winning tickets has been empirically shown on a range of tasks, including computer vision, reinforcement learning, and natural language processing \cite{frankle2018the, yu2019play}. However, the merits of lottery ticket training has recently been called into question. In particular, (1) whether keeping the same initialization (e.g., $\theta_0$) is crucial for acquiring tickets \cite{liu2018rethinking}; and (2) if tickets can generalize across multiple datasets \cite{morcos2019one}.

Our paper investigates the efficacy of lottery tickets when the data distribution changes. We define multiple data domains such that their input distributions are varied. Then, we consider whether subnetworks obtained in a source domain $\mathcal{D}_s$ can be used to specify and train subnetworks in a target domain $\mathcal{D}_t$ where $s \ne t$. Inspired by \citet{liu2018rethinking}, we also experiment with different initialization methods at transfer-time, probing at the importance of initial (source domain) values in disparate target domains. We find that subnetworks obtained through lottery ticket training do not completely overfit to particular input distributions, showing some generalization potential when distributional shifts occur. In addition, we discover a \textit{phase transition} point, at which subnetworks reset to their initial values show better and more stable generalization performance when transferred to an arbitrary target domain.

In summary, our contributions are (1) continuing the line of work on the Lottery Ticket Hypothesis \cite{frankle2018the}, showing that tickets exist in noisy textual domains; (2) performing comprehensive experiments pointing towards the transferability of lottery tickets under distributional shifts in natural language processing; and (3) publicly releasing our code and datasets to promote further discussion on these topics\footnote{https://github.com/facebookresearch/pytext}.

\section{Related Work}
There is a large body of work on transfer learning for neural networks \cite{deng2013sparse, yosinski2014how, liu2017sparse, zoph2018learning, kornblith2019better}. Most of these works focus on improving the transferred representation across tasks and datasets. The representation from a source dataset is fine-tuned or learned collaborately on a target dataset. In contrast, we focus on understanding whether the \textit{architecture} can be transferred and retrained, and whether transferring the initialization is required. Our work is also related to Neural Architecture Search (NAS) \cite{zoph2018learning, liu2018darts, elsken2018neural}. The goal of NAS is to identify well-performing neural networks automatically. Network pruning can be viewed as a form of NAS, where the search space is the sparse topologies within the original over-parameterized network \cite{liu2018rethinking, gale2019state, frankle2018the}.  

Iterative magnitude pruning \cite{frankle2018the, frankle2019stable} is a recently proposed method for finding small, sparse subnetworks from large, over-parameterized neural networks that can be trained in isolation to reach a similar (or better) test accuracy. To obtain these re-trainable sparse subnetworks, \citet{frankle2018the} uses an iterative pipeline that involves training a model, removing ``redundant'' network connections identified by a sparsification algorithm, re-training the subnetwork with the remaining connections. In particular, the experiments in \citet{frankle2018the} show it is critical to re-initialize the subnetworks using the \textit{same} initial values after each round of the iterative pipeline. 

However, the importance of re-using the original initialization is questioned in \citet{liu2018rethinking}, where the authors show that competitive performance of the sparse subnetworks can be achieved with random initialization as well. \citet{morcos2019one} investigate the transferability of lottery tickets across multiple optimizers and datasets for supervised image classification, showing that tickets can indeed generalize \cite{morcos2019one}. Beyond the differences between our domain, task, and datasets, our work carries an important distinction. In \citet{morcos2019one}, the authors refer to the \textit{transfer of initialization} as both the \textit{transfer of the sparse topologies} and the \textit{transfer of the initial values} of the subnetworks. Therefore, it is unclear whether the \textit{sparse topology} alone can be transferred across datasets or the topology combined with the initial values must be exploited jointly to achieve transferability. In our work, we decouple this question by investigating the influence of different initialization strategies on the sparse architecture during the process of finding the winning tickets and after the transfer to other domains.

\section{Task and Datasets}

\paragraph{Distributional Shifts} 
Let $(x^s_i, y^s_i) \in \mathcal{X} \times \mathcal{Y}$ denote a pair of training samples from domain $\mathcal{D}_s$. Let $f(x; \theta)$ be a function (e.g., deep neural network) that maps an input from $\mathcal{X}$ to the label space $\mathcal{Y}$, parameterized by $\theta$. In this work, the sparsity of $\theta$ is induced by the lottery ticket training process \cite{frankle2018the}. To model distributional shifts, we characterize each domain $\mathcal{D}_i$ as a dataset from the Amazon Reviews corpus \cite{mcauley2013hidden}. The differences in unigram frequencies, semantic content, and random noise mimic the type of distributional shifts that occur in machine learning. 

\paragraph{Subword Vocabulary} We ensure each domain $\mathcal{D}$ shares an identical support on $\mathcal{X}$ by encoding the inputs using a vocabulary common across all datasets. Word-level vocabularies may introduce problems during domain transfer as certain words potentially only appear within a particular domain. On the other end of the spectrum, character-level vocabularies ameliorate this issue but may not contain enough expressive power to model the data. We elect to use a subword vocabulary, balancing the out-of-vocabulary and effectiveness problems introduced by the word- and character-level vocabularies, respectively. Technical details for creating the shared subword vocabulary are presented in \S\ref{method:vocab}.

\begin{figure}[t]
    \centering
    \includegraphics[scale=0.4]{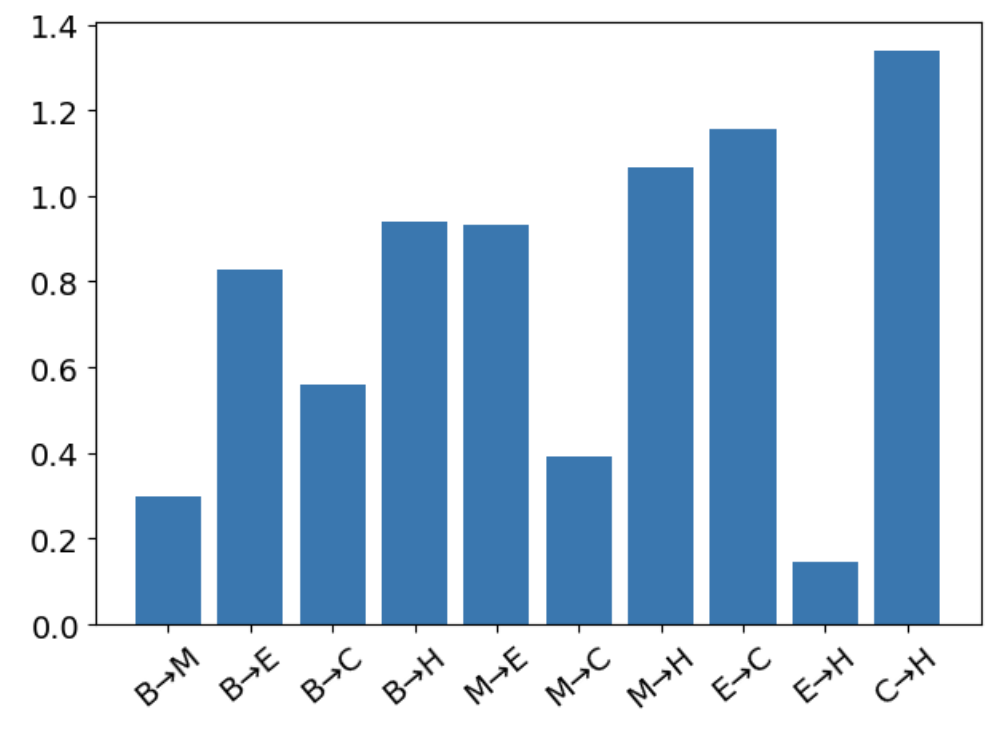}
    \caption{Jenson-Shannon Divergence scores on subword unigram distributions for each domain pair $(\mathcal{D}_i, \mathcal{D}_{i'})$. Domains include Books (B), Electronics (E), Movies (M), CDs (C), and Home (H). Values are scaled by $1e^5$ for presentation.}
    \label{fig:jsd}
\end{figure}

\paragraph{Divergence Scores} Given an identical support for all data distributions, we now quantify the distributional shifts between our domains using Jenson-Shannon Divergence (JSD). JSD is a symmetric measure of similarity between two (continuous) probability distributions $p$ and $q$ with a proxy, averaged distribution $m = \frac{1}{2}(p+q)$:
\begin{equation}
    \mathrm{JSD}(p||q) = \frac{1}{2}\mathrm{KL}(p||m) + \frac{1}{2}\mathrm{KL}(q||m)
    \label{eq:jsd}
\end{equation}
where $\mathrm{KL}(p||q)$ in Eq. \ref{eq:jsd} denotes the Kullback-Leibler divergence, defined as:
\begin{equation}
    \mathrm{KL}(p||q) = \int_{-\infty}^{\infty} p(x)\log \frac{p(x)}{q(x)}dx
\end{equation}

Figure \ref{fig:jsd} displays the divergence scores between our datasets. On average, there is high disagreement with respect to the prevalence and usage of subwords in each domain, with Electronics$\rightarrow$Home the most similar and CDs$\rightarrow$Home the most dissimilar.

\paragraph{Sentiment Analysis} Finally, we introduce our base task for experimentation. Our models are evaluated on a binary sentiment analysis task constructed from five categories in the Amazon Reviews corpus: books (B), electronics (E), movies (M), CDs (C), and home (H). The dataset originally provides fine-grained sentiment labels ($1$ through $5$) so we group $1$, $2$ as negative and $4$, $5$ as positive. Following \citet{peng2018cross}, reviews with neutral ratings ($3$) are discarded. We sample 20K train, 10K validation, and 10K test samples from each category, ensuring there is an equal distribution of positive and negative reviews.

\section{Methods}

In this section, we discuss our technical methods. First, we describe the subword vocabulary creation process (\S\ref{method:vocab}). Second, we cover the underlying model used in the sentiment analysis task (\S\ref{method:model}). Third, we detail the lottery ticket training and transferring methods (\S\ref{method:tickets}).

\subsection{Vocabulary} \label{method:vocab}

We use the SentencePiece\footnote{https://github.com/google/sentencepiece} library to create a joint subword vocabulary for our datasets \cite{kudo2018sentencepiece}. The subword model is trained on the concatenation of all five training datasets (100K sentences) using the byte-pair encoding algorithm \cite{sennrich2016neural}. We set the vocabulary size to 8K. The final character coverage is 0.9995, ensuring minimal out-of-vocabulary problems during domain transfer. 

\subsection{Model} \label{method:model}

We use convolutional networks (CNN) as the underlying model given their strong performance on numerous text classification tasks \cite{kim2014convolutional, mou2016transferable, gehring2017convolutional}. Let $V$ and $n$ represent the vocabulary of the corpus and maximum sequence length, respectively. Sentences are encoded as an integer sequence $t_1, \cdots, t_n$ where $t_i \in V$. The embedding layer replaces each token $t_i$ with a vector $\mathbf{t}_i \in \mathbb{R}^d$ that serves as the corresponding $d$-dimensional embedding. The vectors $\mathbf{t}_1, \cdots, \mathbf{t}_n$ are concatenated row-wise to form a token embedding matrix $\mathbf{T} \in \mathbb{R}^{n \times d}$.

Our model ingests the embedding matrix $\mathbf{T}$, then performs a series of convolutions to extract salient features from the input.  We define a convolutional filter $\mathbf{W} \in \mathbb{R}^{h \times d}$ where $h$ represents the \textit{height} of the filter. The filter is not strided, padded, or dilated, Let $\mathbf{T}[i:j] \in \mathbb{R}^{h\times d}$ represent a sub-matrix of $\mathbf{T}$ extracted from rows $i$ through $j$, inclusive. The feature map $\mathbf{c} \in \mathbb{R}^{n-h+1}$ is induced by applying the filter to each possible window of $h$ words, i.e.,
\begin{equation}
    c_i = f\Big( \big\langle \mathbf{T}[i:i+h],\mathbf{W}\big\rangle_{\fro} + b\Big) 
\end{equation}
for $1\leq i \leq n-h+1$, where $b \in \mathbb{R}$ is a bias term, $f$ is a non-linear function, and the Frobenius inner product is denoted by $\langle \mathbf{A},\mathbf{B}\rangle_{\fro} = \sum_{i=1}^h \sum_{j=1}^d \mathbf{A}_{ij} \mathbf{B}_{ij}$. 1-max pooling \cite{collobert2011natural} is applied on $\mathbf{c}$, defined as $\hat{c} = \textnormal{max}\{\mathbf{c}\}$. This is performed to propagate the maximum signal throughout the network and reduce the dimensionality of the input.

The process described above creates \textit{one} feature from \textit{one} convolution with window $h$ followed by a pooling operation. To extract multiple features, the model uses several convolutions with varying $h$ to obtain features from different sized $n$-grams in the sequence. The convolutional (and pooled) outputs are concatenated along the channel dimension, then fed into a one-layer MLP to obtain a distribution over the $c$ classes.

\subsection{Lottery Tickets} \label{method:tickets}

\subsubsection{Initialization}
\label{sec:init}
The embedding matrix is initialized from a unit Gaussian, $\mathbf{T} \sim \mathcal{N}(0,1)$. The convolutional and MLP layers use He initialization \cite{he2015delving}, whose bound is defined as
\begin{equation}
    b = \sqrt{\frac{6}{(1+a^2) \times \mathrm{fan\_in}}}
\end{equation}
where $a$ and $\mathrm{fan\_in}$ are parameters calculated for each weight. The resulting weights have values uniformly sampled from $\mathcal{U}(-b,b)$.

\subsubsection{Training}
\label{sec:train}
We use iterative pruning with alternating cycles of training and pruning to obtain the tickets \cite{han2015learning, frankle2018the}. 
For clarity, we define a \textit{round} as training a network for a fixed number of epochs. We begin with a seed round $r_0$ where the model does not undergo any pruning, then begin to procure tickets in a series of lottery ticket training rounds.

In each successive round $r_{i>0}$, a fraction $p$ of the weights that survived round $r_{i-1}$ are pruned (according to a sparsification algorithm, discussed below) to obtain a smaller, sparser subnetwork; this is denoted by $f(x;m_i\odot \theta_i)$ where $m_i$ and $\theta_i$ represent the sparse mask and weights at round $r_i$. The weights $\theta_i$ of this subnetwork are set according to an \textit{initialization strategy} and the subnetwork is re-trained to convergence. We refer to the \textit{sparsity} as the fraction of weights in the network that are exactly zero. In each round, we prune $p\%$ of the weights in the model. Therefore, the resulting ticket has sparsity $1-(1 - p\%)^{r_{total}}$, where $r_{total}$ is the total number of lottery ticket training rounds. 

Next, we discuss the sparsification algorithm used to prune weights in each round $r_i$. Let $\mathbf{p}_i$ denote the vectorized collection of trainable parameters in layer $i \ge 0$, with the embedding layer as layer $0$. After re-training the (sub-)networks in each round, we apply the $\ell_0$ projection on the parameters in each layer, i.e.
\begin{equation}
    \argmin_{\mathbf{p}} ||\mathbf{p}-\mathbf{p}_i||^{2}_{2}
    \label{eq:10}
\end{equation}
subject to $\card(\mathbf{p}) \le k_i$, where $\card(\mathbf{p})$ denotes the number of non-zeros in $\mathbf{p}$. The optimization problem in Eq. \ref{eq:10} can be solved analytically by sorting the elements of $\mathbf{p}_i$ with respect to their absolute values and picking the top $k_i$ elements with the largest magnitude \cite{jain2017non, zhu2017prune}.  We use the sparsity hyperparameter $p$ introduced above to decide $k_i$ for each layer. Let $\mathrm{len}(\mathbf{p}_i)$ denote the total number of trainable parameters in layer $i$. We set $k_i = p\% \times \mathrm{len}(\mathbf{p}_i)$ for each layer. In accordance with our training procedure, once a weight is pruned, it is no longer a trainable parameter; hence, $\mathrm{len}(\mathbf{p}_i)$ is strictly decreasing after each round.

\begin{figure}[t]
    \centering
    \includegraphics[scale=0.15]{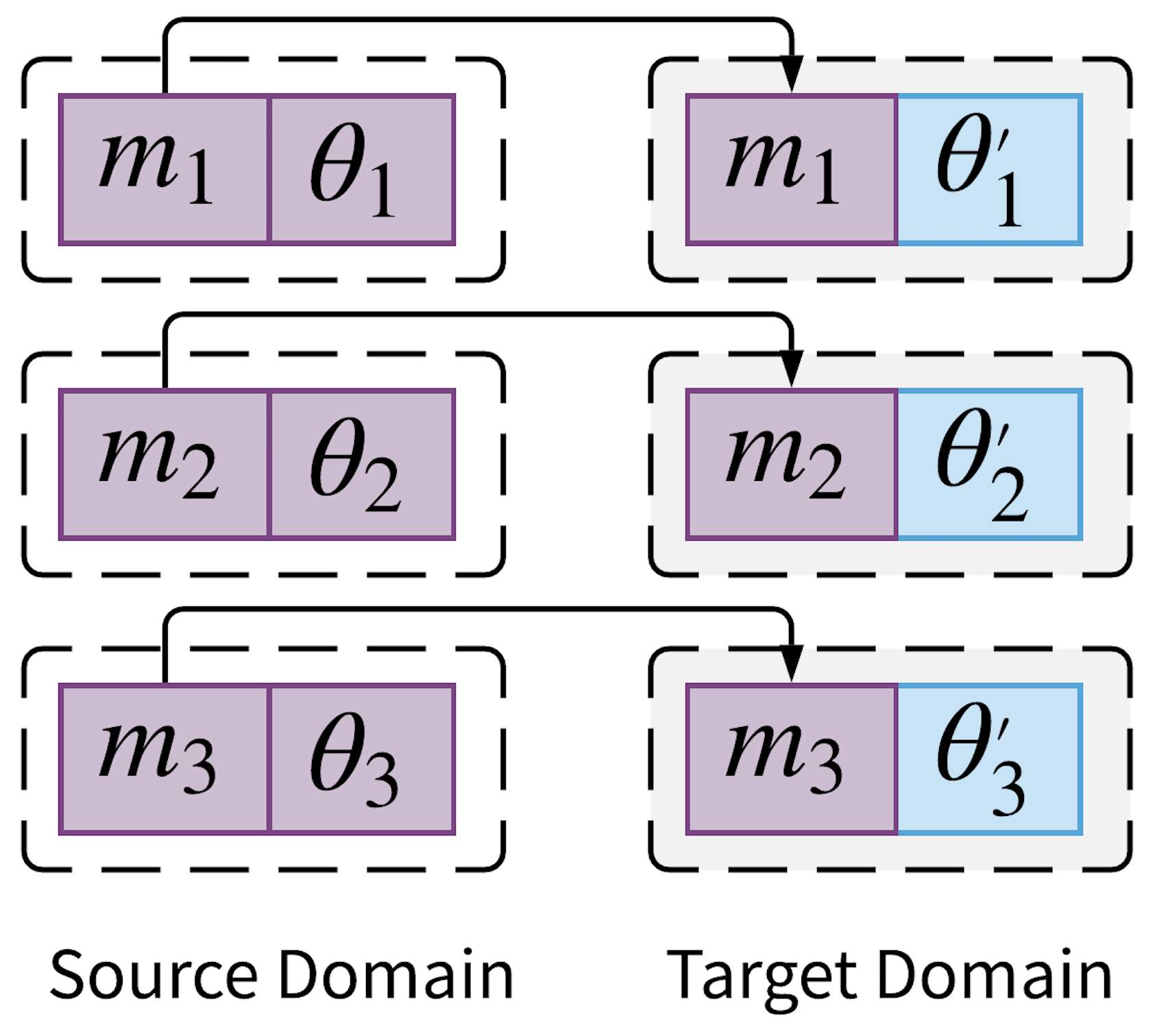}
    \caption{Visualization of the subnetwork transfer process. Purple denotes elements from the source domain, while blue denotes elements from the target domain. Tickets are composed of two elements: (1) the sparsified mask ($m_i$) and (2) the initial parameter values ($\theta_i$). During transfer, we create subnetworks in the source domain with the mask borrowed from the source domain, but with potentially different parameters. We use $\theta_i'$ to denote that these parameters are set according to some \textit{initialization strategy}, which we discuss further in our experiments (\S\ref{sec:exp}).}
    \label{fig:clarify}
\end{figure}

\begin{figure*}[ht!]
    \begin{center}
    \includegraphics[width=0.325\textwidth]{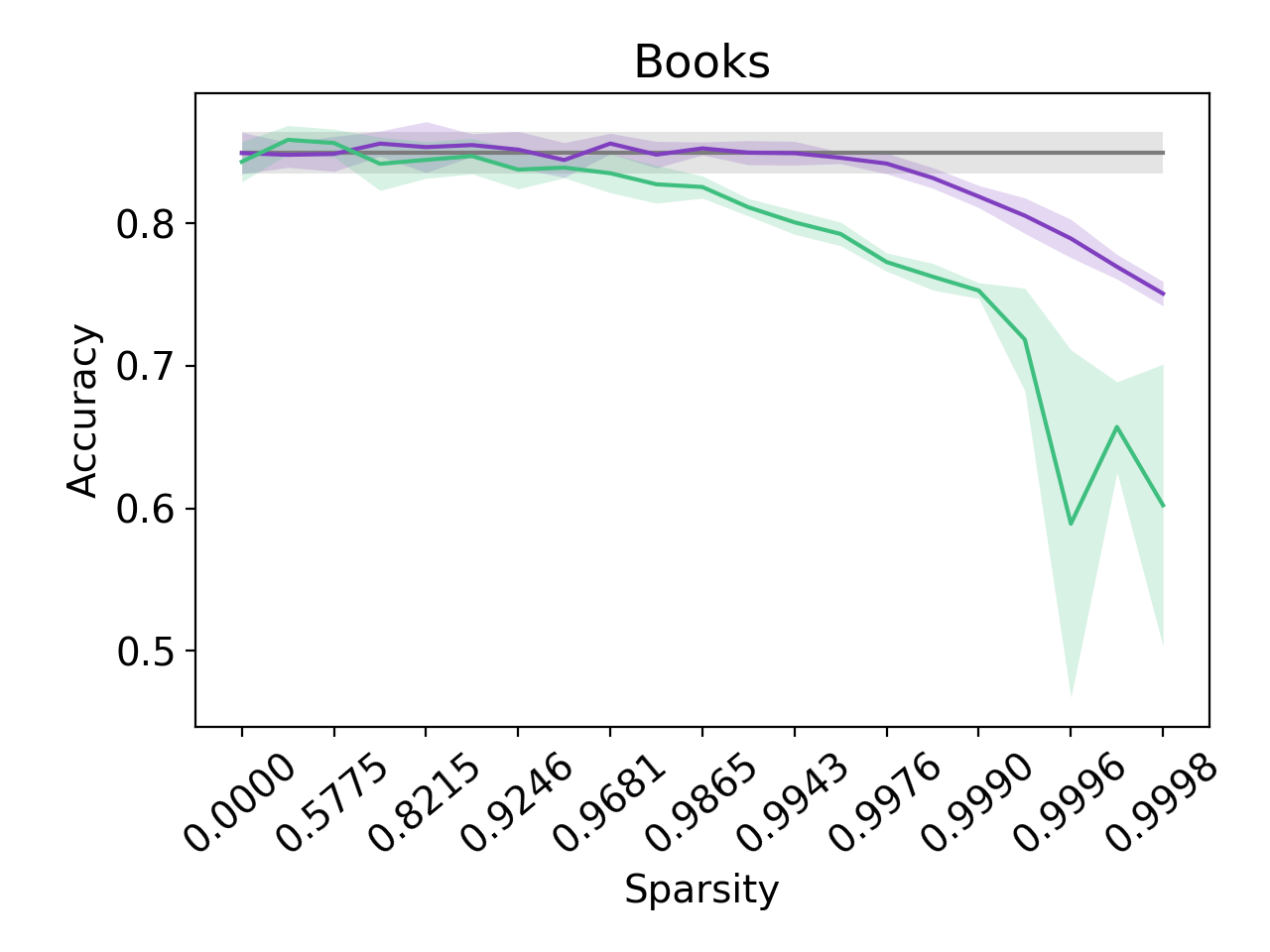}
    \includegraphics[width=0.325\textwidth]{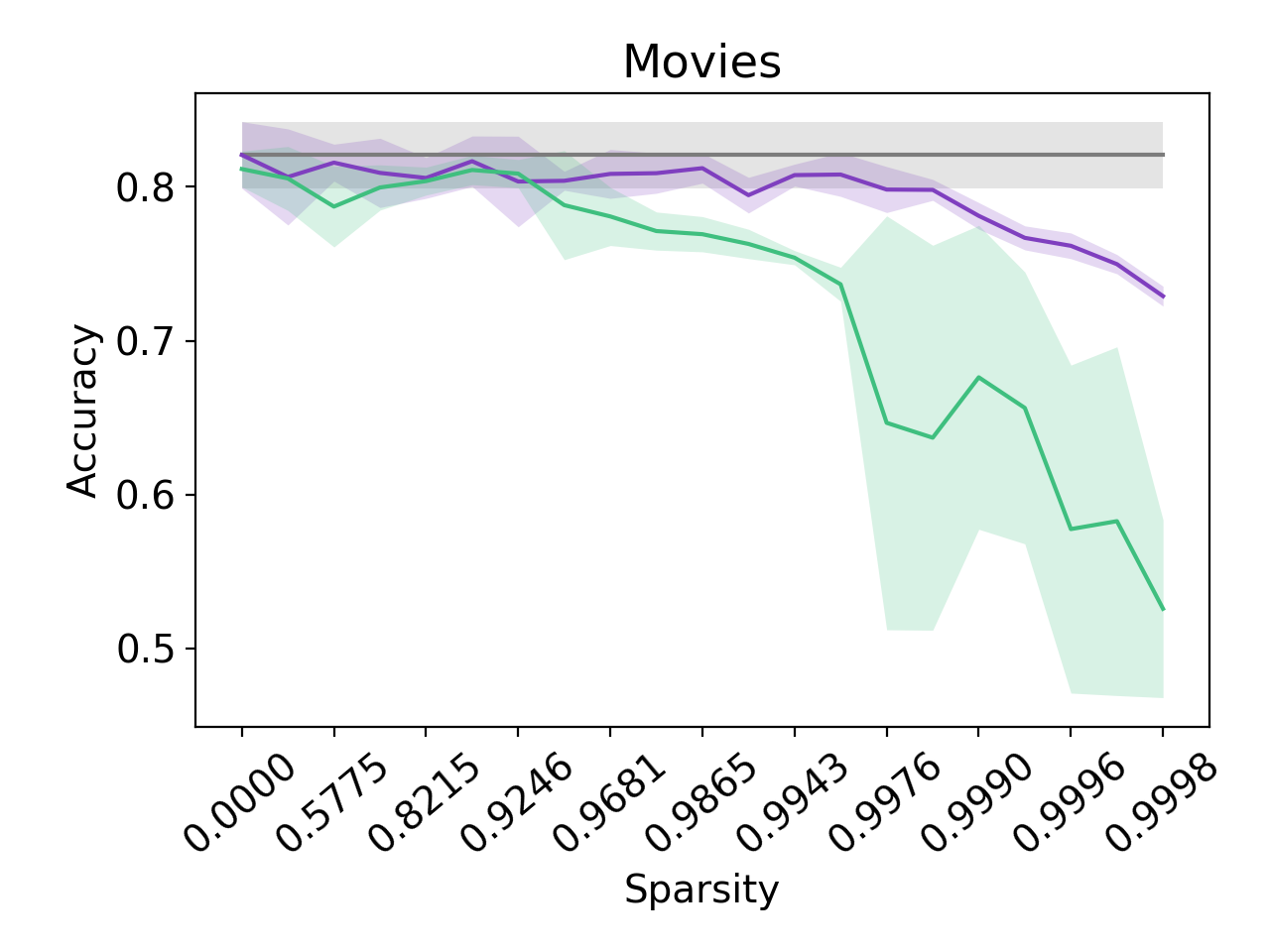}
    \includegraphics[width=0.325\textwidth]{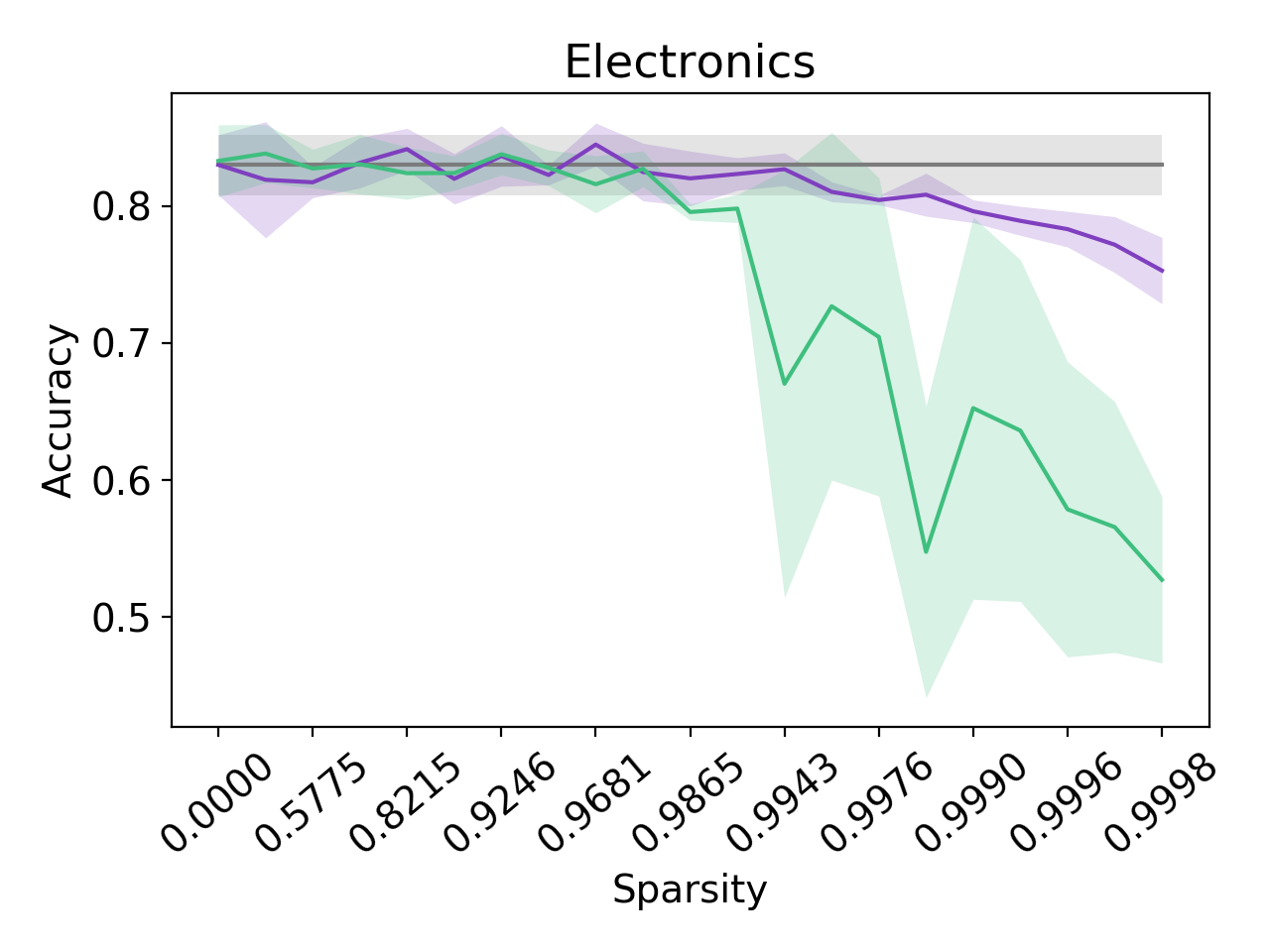}
    \includegraphics[width=0.325\textwidth]{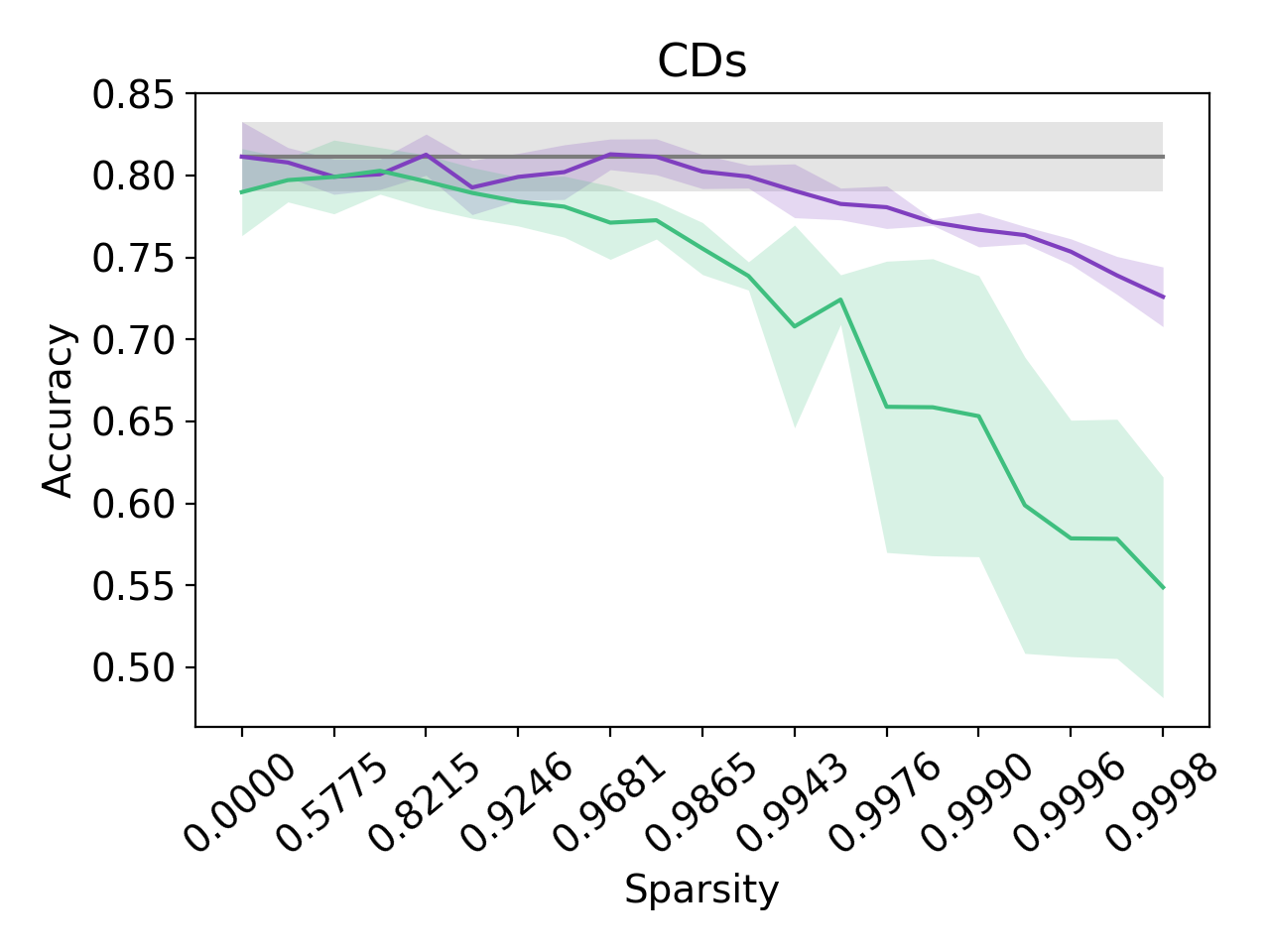}
    \includegraphics[width=0.325\textwidth]{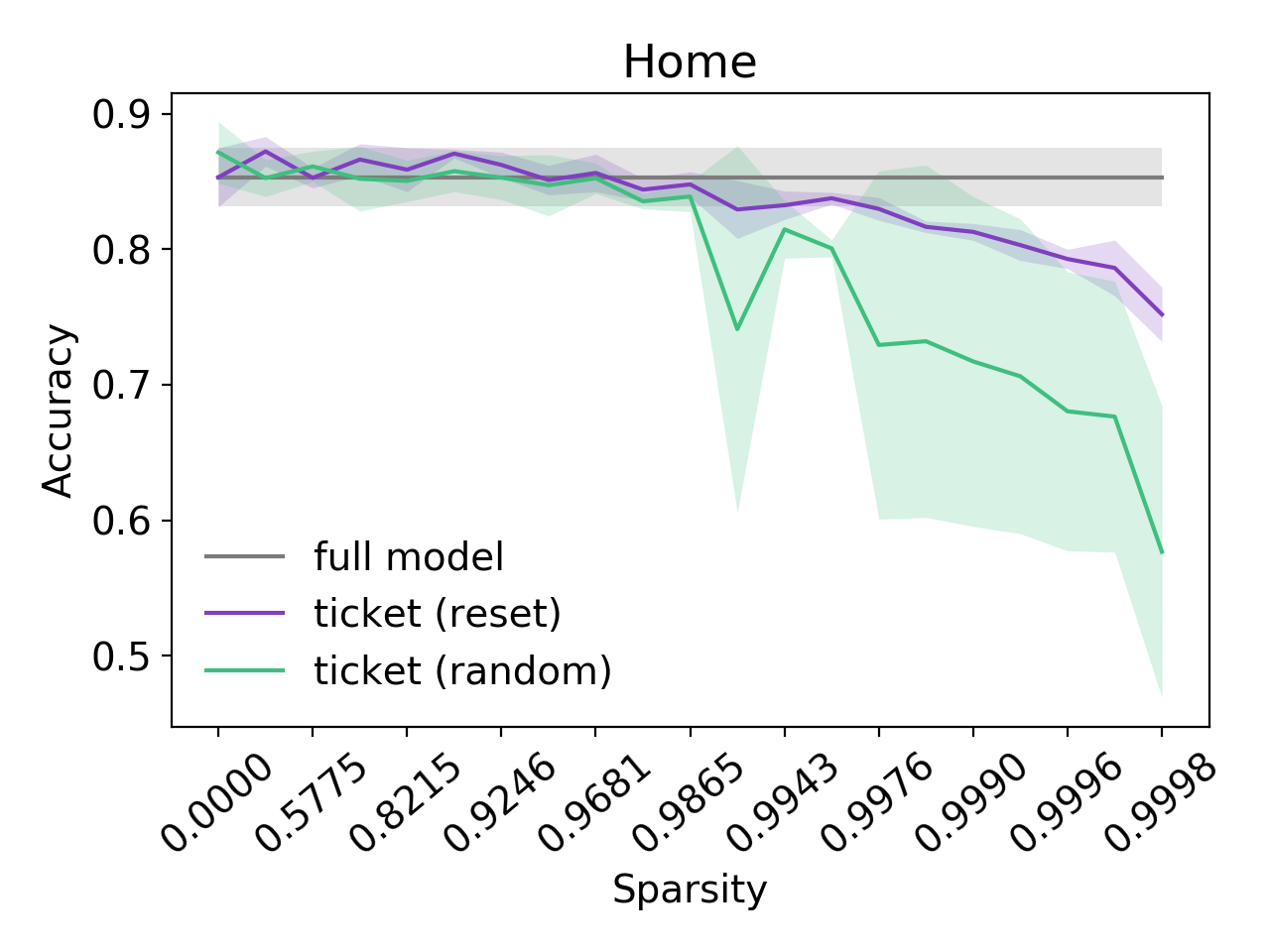}
    \end{center}
    \label{fig:books}
    \caption{Results obtaining lottery tickets on the Books, Movies, Electronics, CDs, and Home categories of the Amazon Reviews dataset \cite{mcauley2013hidden}. Experiments are repeated five times, where the solid lines represent the mean and shaded regions represent the standard deviation. Note that the $x$-axis ticks are \textit{not} uniformly spaced.}
    \label{fig:obtain}
\end{figure*}

\begin{figure*}[t]
    \begin{center}
    \includegraphics[width=0.325\textwidth]{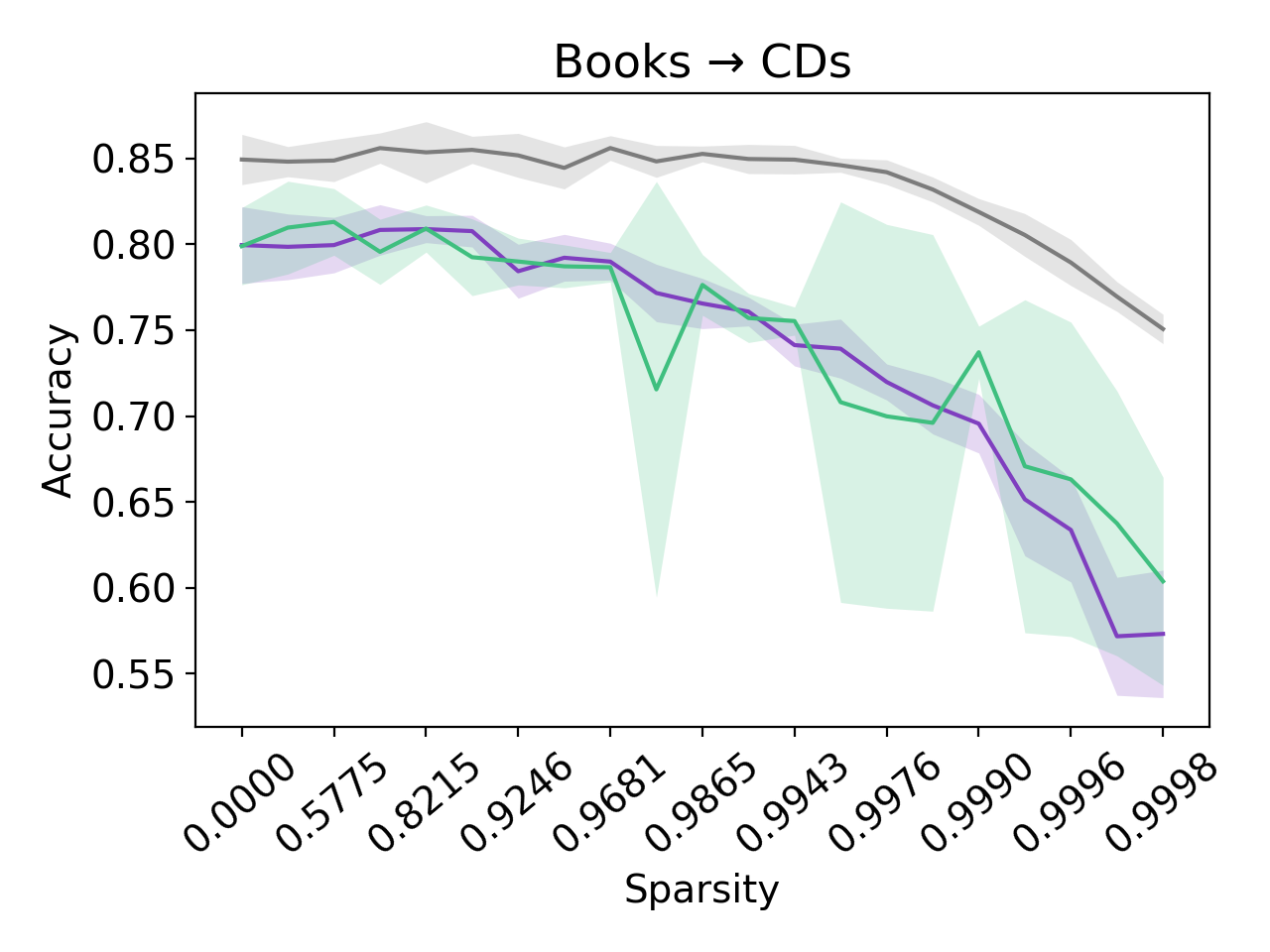}
    \includegraphics[width=0.325\textwidth]{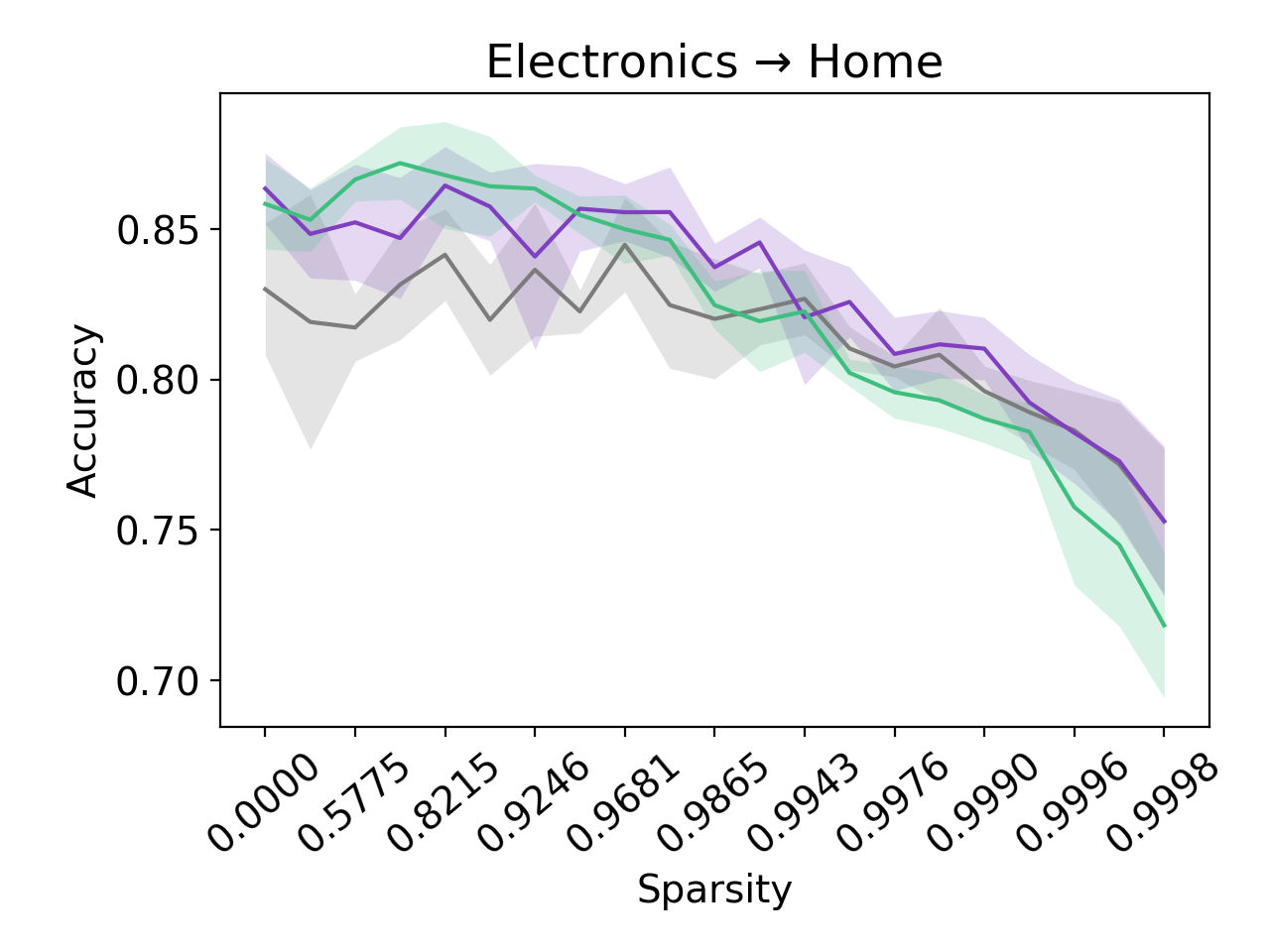}
    \includegraphics[width=0.325\textwidth]{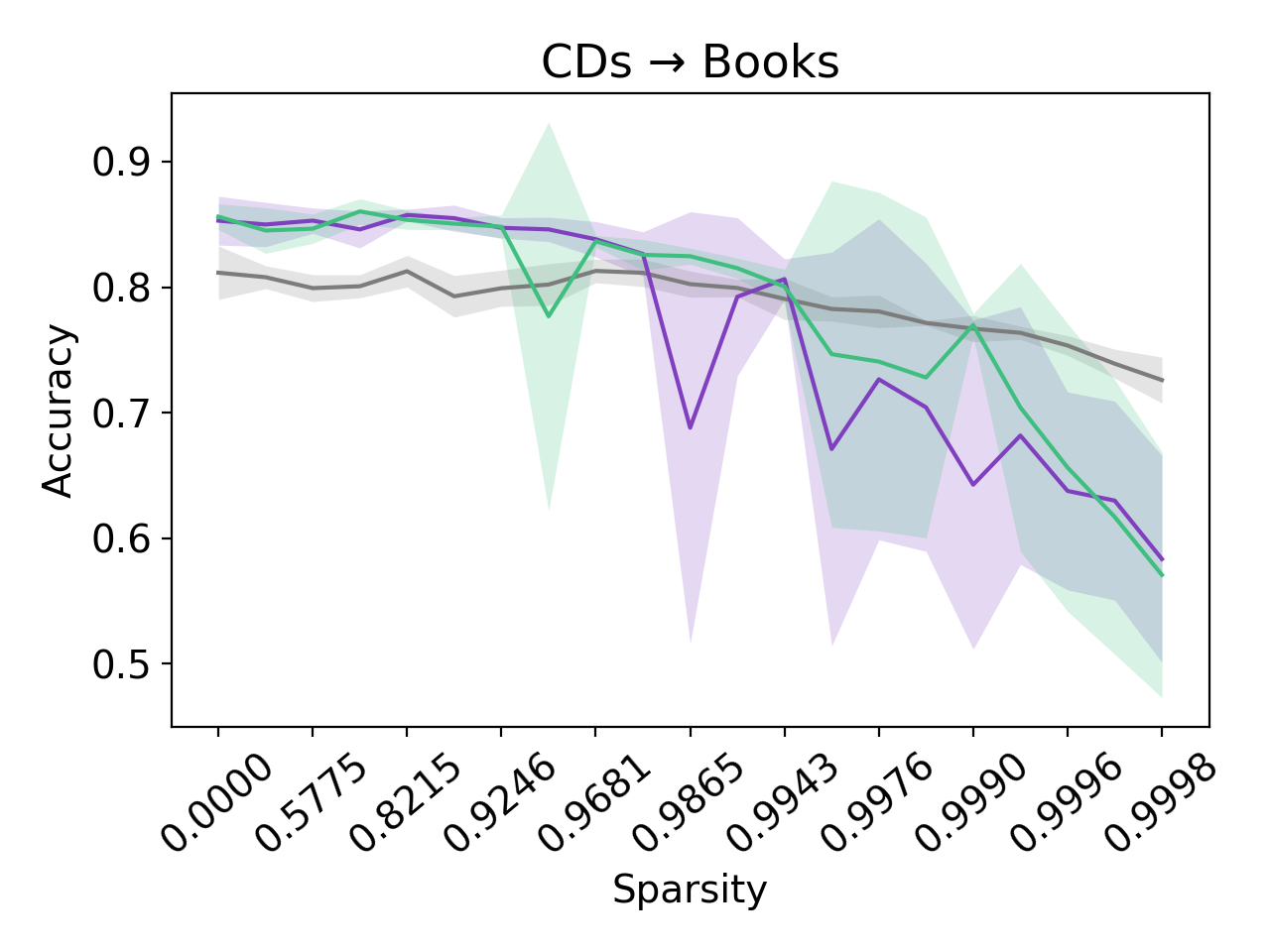}
    \includegraphics[width=0.325\textwidth]{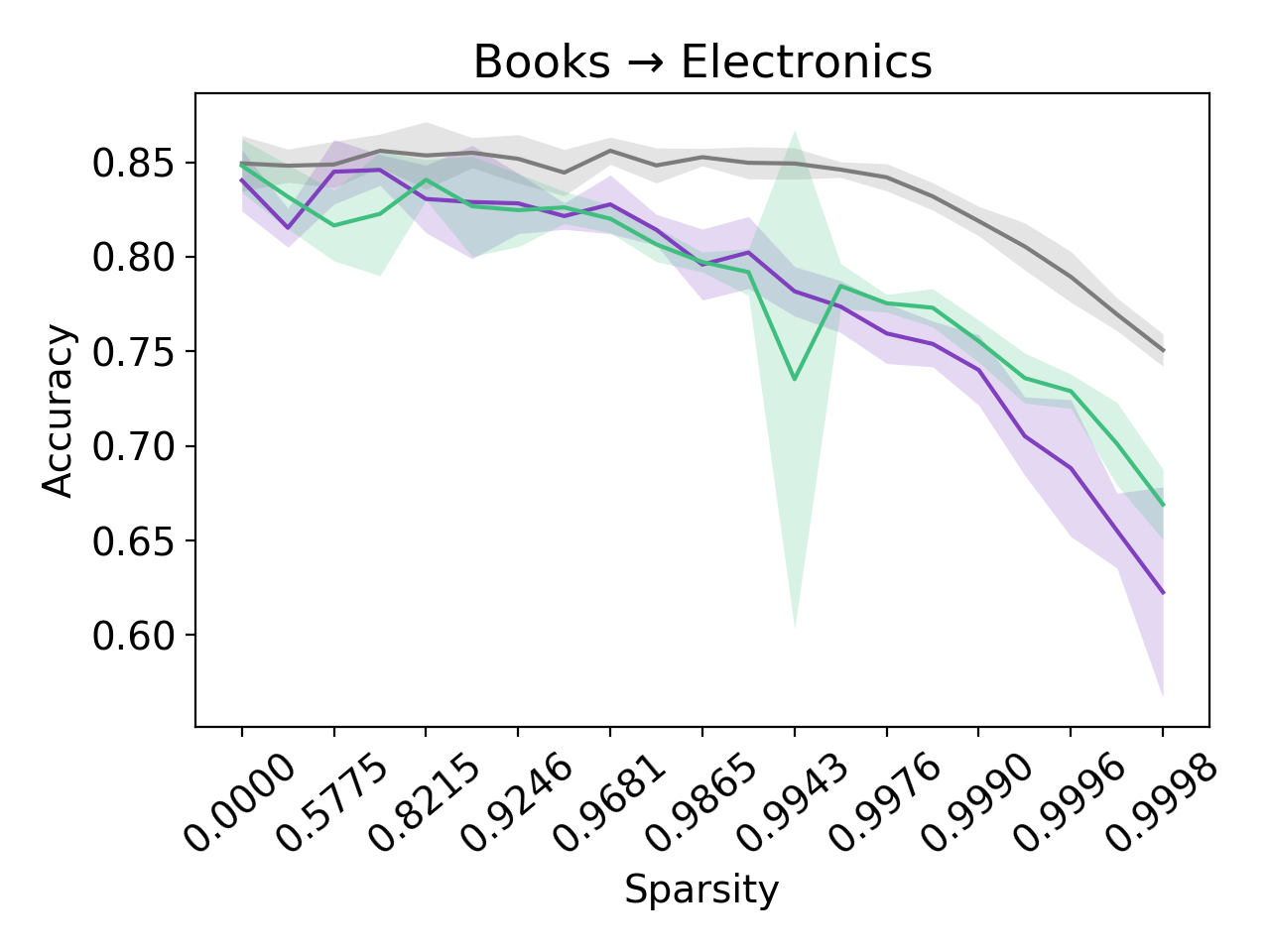}
    \includegraphics[width=0.325\textwidth]{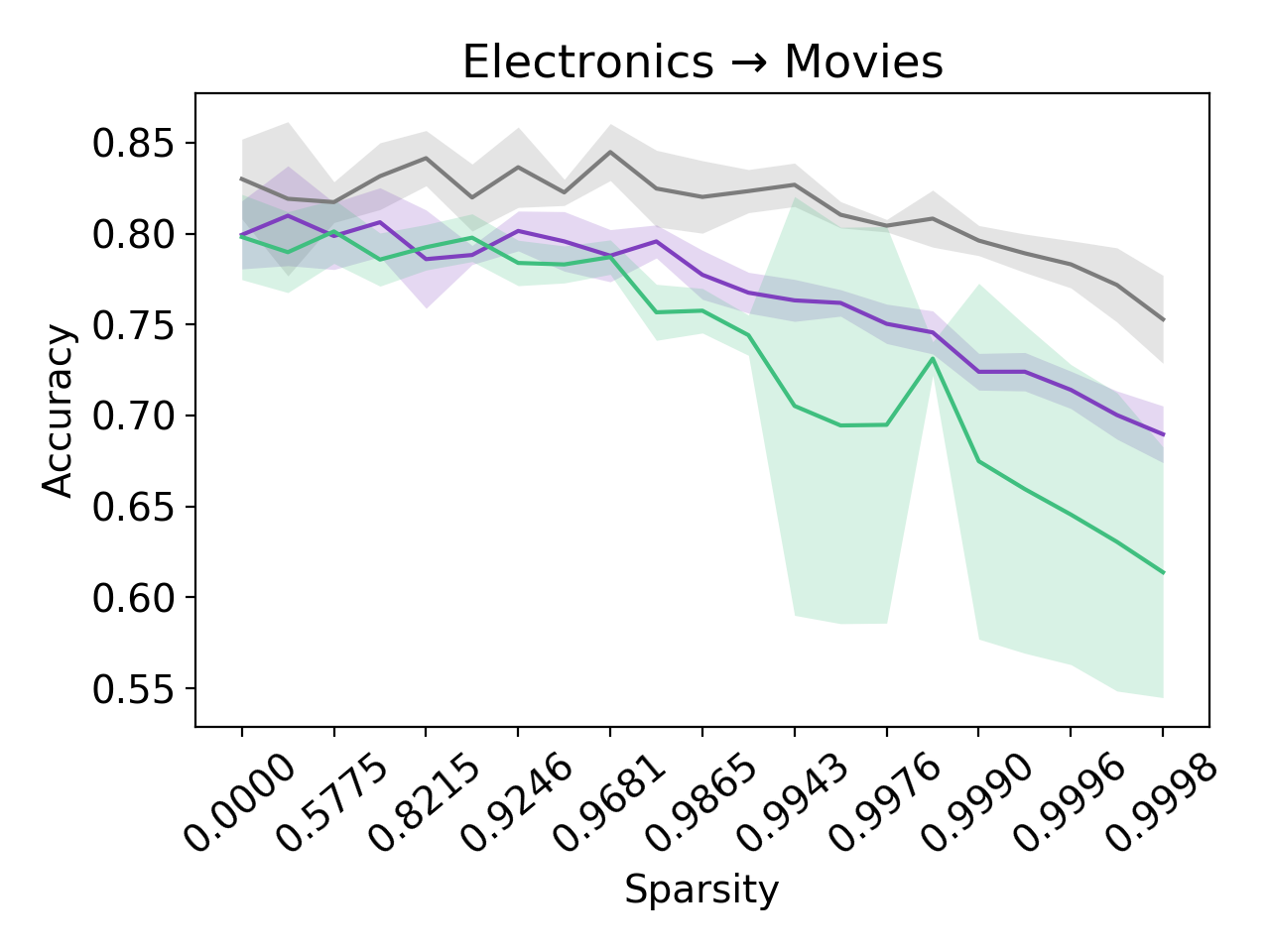}
    \includegraphics[width=0.325\textwidth]{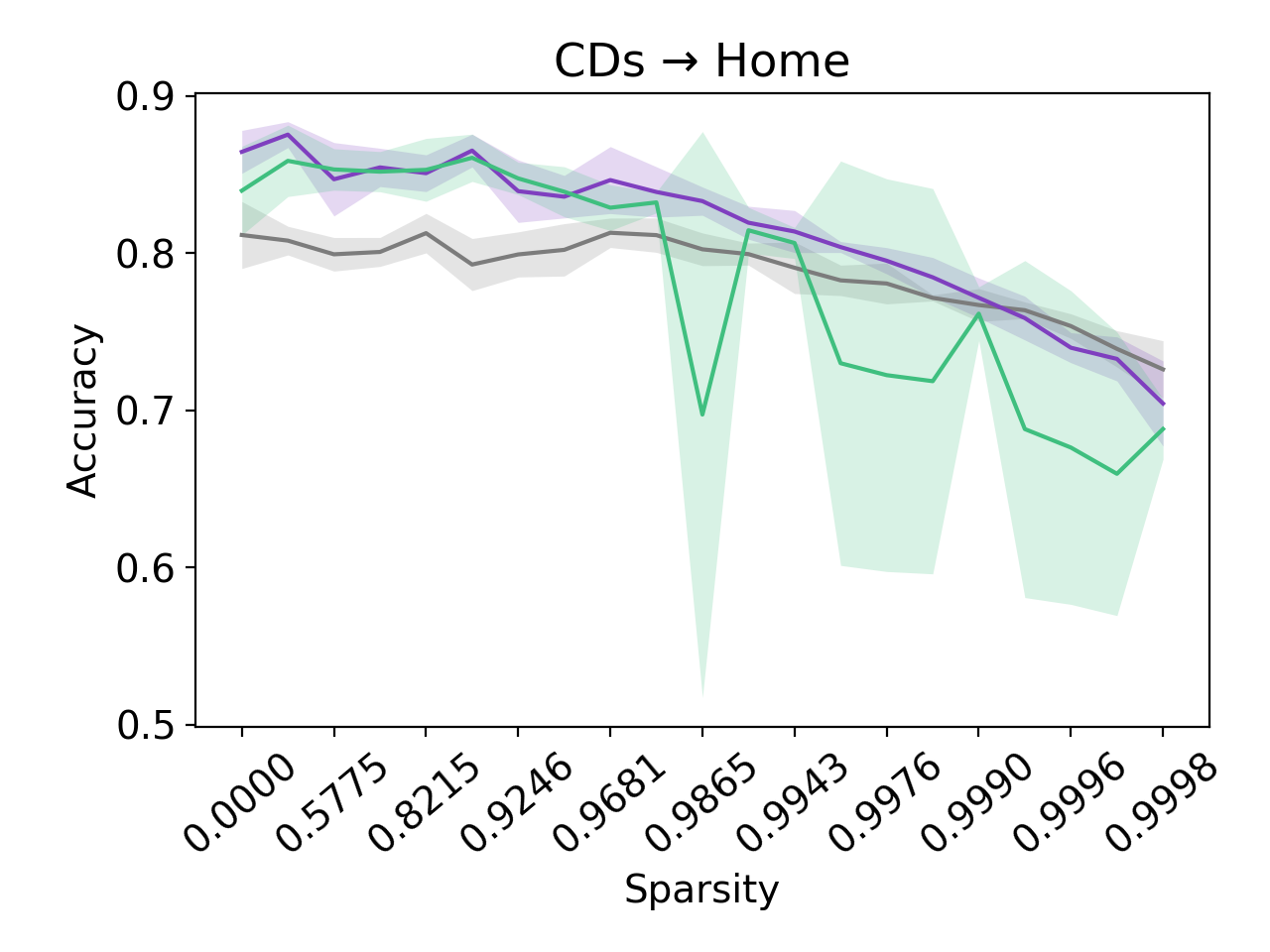}
    \includegraphics[width=0.325\textwidth]{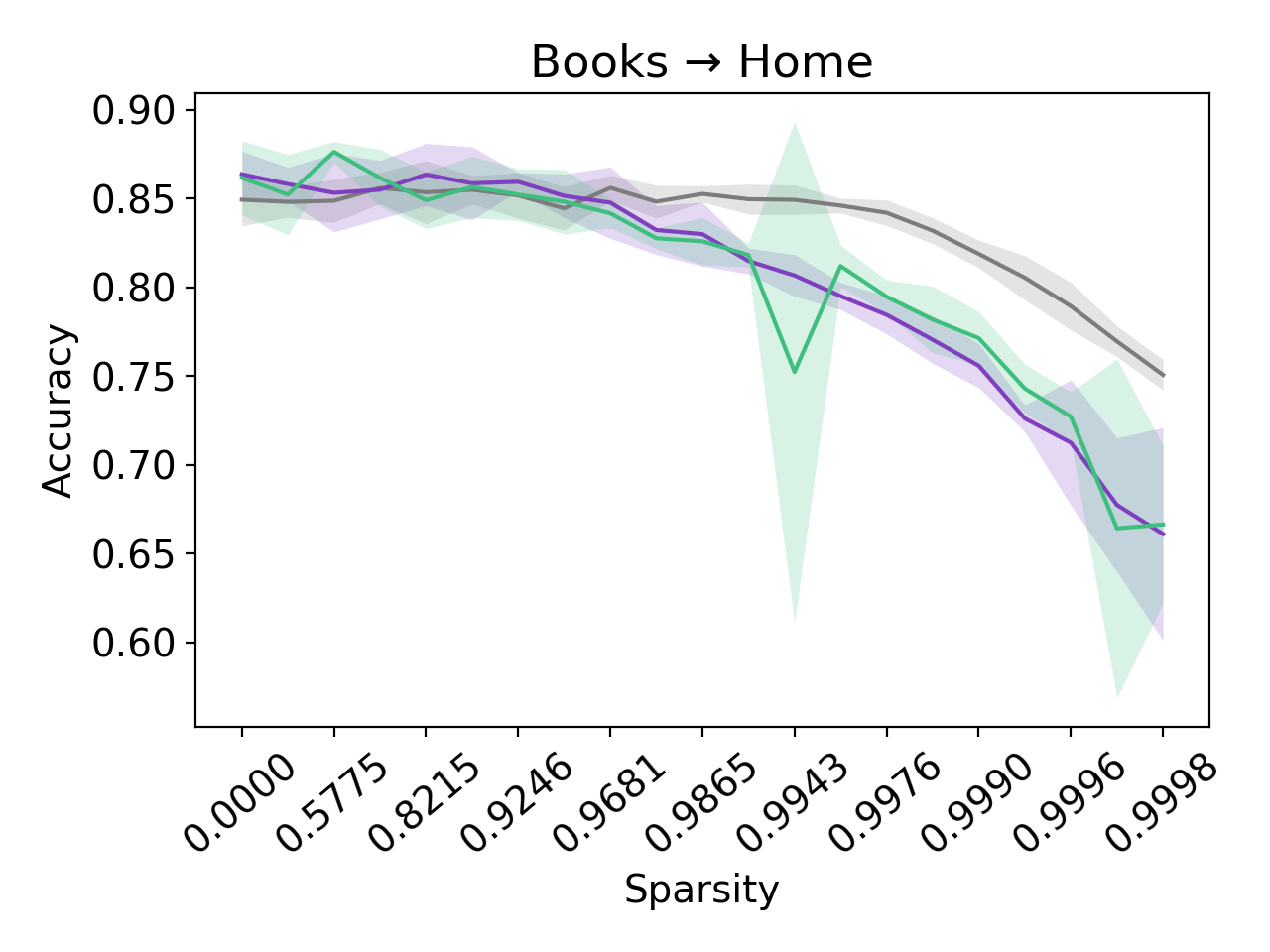}
    \includegraphics[width=0.325\textwidth]{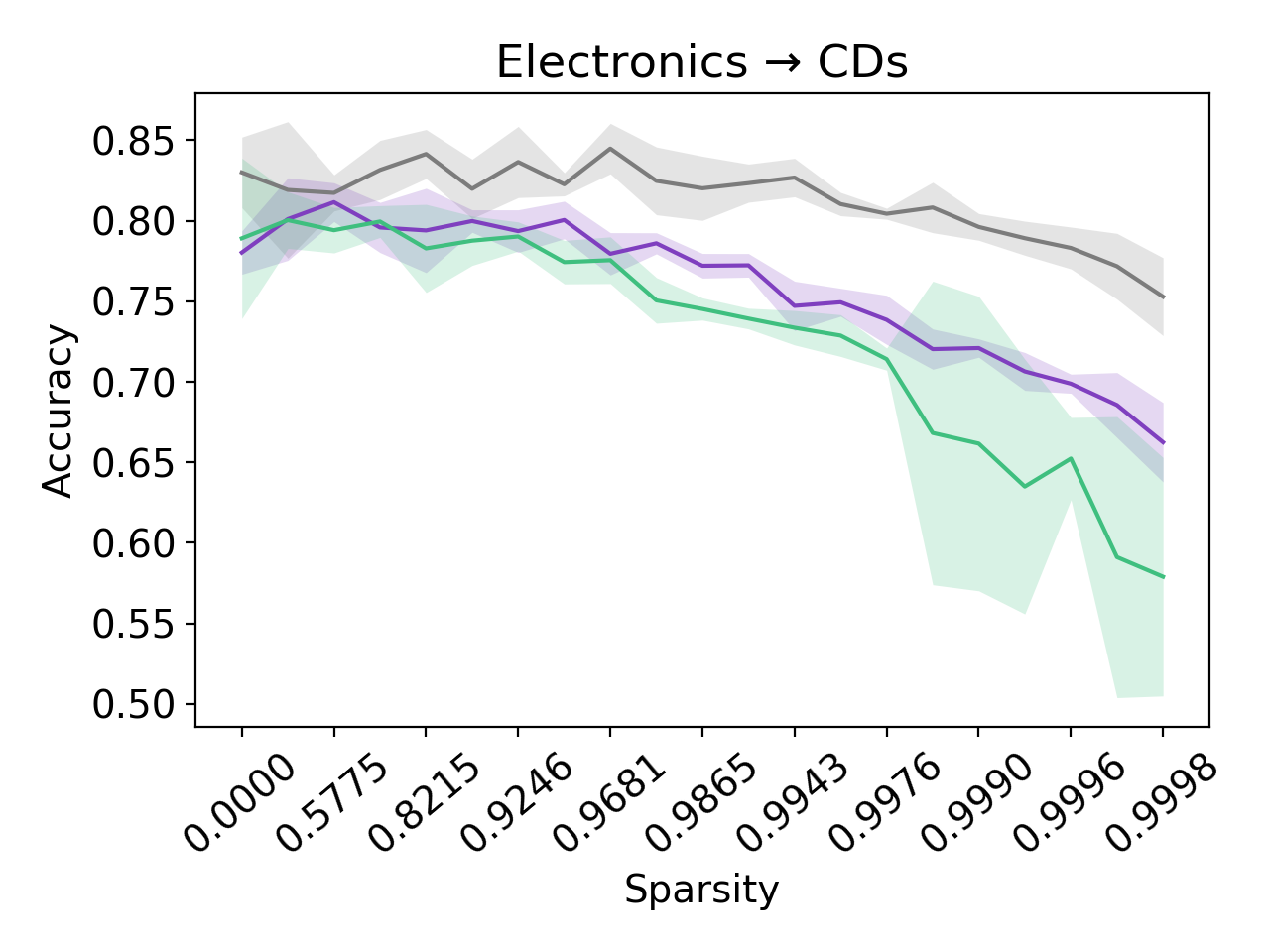}
    \includegraphics[width=0.325\textwidth]{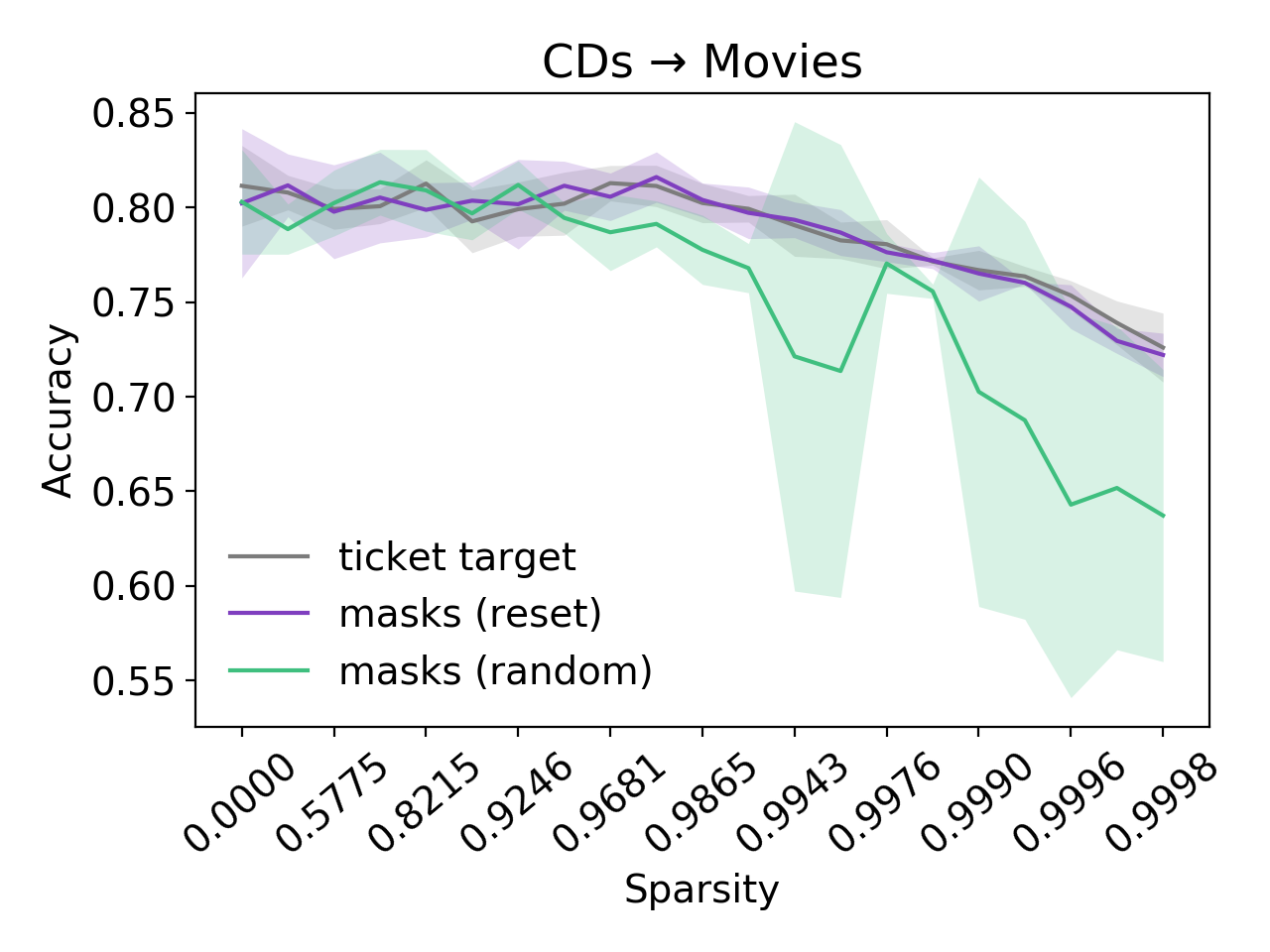}
    \end{center}
    \label{fig:books}
    \caption{Results transferring lottery tickets on nine transfer tasks constructed from the five categories of the Amazon Reviews dataset \cite{mcauley2013hidden}. Experiments are repeated five times, where the solid lines represent the mean and shaded regions represent the standard deviation. Note that the $x$-axis ticks are \textit{not} uniformly spaced.}
    \label{fig:transfer}
\end{figure*}

\subsubsection{Transferring}

The lottery ticket training procedure outlined in \S\ref{sec:train} yields a batch of subnetworks $f(x^s;m_1\odot \theta), \cdots, f(x^s;m_n\odot \theta)$ where $x^s$ represents the inputs from a \textit{source} domain $\mathcal{D}_s$ and $m_i$ represents the sparse mask used to prune weights at round $r_i$. During transfer, we construct a new batch of subnetworks $f(x^t;m_1\odot \theta'), \cdots, f(x^t;m_n\odot \theta')$ to be evaluated on inputs from a (non-identical) \textit{target} domain $\mathcal{D}_t$ with masks derived from the \textit{source} domain. The change in parameter notation ($\theta \rightarrow \theta'$) implies that the subnetworks evaluated in a disparate domain can potentially use a different \textit{transfer} initialization strategy. We clarify this process in Figure \ref{fig:clarify}. In contrast, \citet{morcos2019one} transfers the entire ticket (sparse masks and initial values) to the target domain. Finally, using the new batch of subnetworks, we evaluate each subnetwork $f(x^t;m_i \odot \theta')$ in the target domain for $r_{total}$ rounds. Unlike the canonical ticket training rounds, we do not (additionally) sparsify the subnetworks during transfer. All in all, our transfer task is designed to answer the following question: can the \textit{sparse masks} found in a source domain using lottery ticket training (\S\ref{method:tickets}) be transferred to a target domain with \textit{different initialization strategies} to match the performance of a ticket obtained in same target domain?

\section{Experiments} \label{sec:exp}
\subsection{Settings}
\label{sec:expsetting}
Our CNN uses three filters ($h \in [3,4,5]$), each with $127$ channels, and ReLU activation \cite{nair2010rectified}. We fix the maximum sequence length to $500$ subwords. The embeddings are $417$-dimensional and trained alongside the model. We opt not to use pre-trained embeddings to ensure the generalizability of our results. Additionally, we regularize the embeddings with dropout \cite{srivastava2014dropout}, $p=0.285$. The MLP contains one hidden layer with a dimension of $117$. Hyperparameters were discovered using Bayesian hyperparameter optimization \cite{snoek2012practical} on the Books validation set. The models are trained with a batch size of $32$ for a maximum of $15$ epochs. Early stopping is used to save iterative model versions that perform well on a development set. We use the Adam optimizer \cite{kingma2014adam} with a learning rate of $1e^{-3}$ and $\ell_2$ regularization with a weight of $1e^{-5}$.

\subsection{Obtaining Tickets} \label{exp:obtain}

First, we use the lottery ticket training procedure outlined in \S\ref{sec:train} to obtain tickets for our five datasets with $p=35\%$ and $r_{total}=20$. We compare the test performance of the subnetworks using the following baselines:
\begin{itemize}
    \item \textsc{Full-Model:} This baseline evaluates the performance of the original network \textit{without} any pruning. In other words, we train a model for a seed round $r_0$, then record its performance.
    \item \textsc{Ticket-Reset:} The values of the subnetwork are reset to their \textit{original values} before training. This initialization strategy was used in the earliest formation of the Lottery Ticket Hypothesis \cite{frankle2018the}.
    \item \textsc{Ticket-Random:} The values of the subnetwork are reset to \textit{random values} drawn from the initialization distribution(s) of the original network. We sample weights from the distributions outlined in \S\ref{sec:init} to initialize the subnetworks.
\end{itemize}

The results are shown in Figure \ref{fig:obtain}. For all datasets, \textsc{Ticket-Reset} shows the best performance, notably outperforming \textsc{Full-Model} in early stages of sparsification (0-90\%) for the Books, Electronics, and Home datasets. This demonstrates that deep neural networks---especially those for sentiment analysis---are highly over-parameterized, and the sparsity induced by lottery ticket training can help to increase performance. This observation is consistent with \citet{louizos2018learning}, which also showed sparse networks fashion a regularization effect that results in better generalization performance. In addition, we observe that \textsc{Ticket-Reset} and \textsc{Ticket-Random} have similar test performance until about 96\% sparsity. This casts some doubt around whether the initial values truly matter for sparse models as the randomly sampled values seem to fit sparse masks well.

However, a \textit{phase transition} occurs in the high sparsity regime, where the differences between \textsc{Ticket-Reset} and \textsc{Ticket-Random} are significantly enlarged. The performance of \textsc{Ticket-Random} becomes highly unstable and drops off much faster than \textsc{Ticket-Reset} after 96\% sparsity. In contrast, \textsc{Ticket-Reset} remains relatively stable---even with sparsity levels over 99.9\%---pointing towards the enigmatic importance of original values in extreme levels of sparsity.

\subsection{Transferring Tickets}

Next, we use the lottery ticket transferring procedure outlined in \S\ref{method:tickets} to transfer (obtained) subnetworks from a \textit{source} domain to a non-identical \textit{target} domain. Identical to the previous experiment, we use $r_{total}=20$. We compare the test performance of the \textit{transferred} subnetworks using the following baselines:
\begin{itemize}
    \item \textsc{Ticket-Target:} This baseline is comprised of the subnetworks obtained in the target domain using lottery ticket training. We borrow the values for this baseline (without modification) from the \textsc{Ticket-Reset} subnetworks shown in Figure \ref{fig:obtain}, albeit from the domain of interest.
    \item \textsc{Masks-Reset:} Under this initialization strategy, the masks obtained in the source domain is used on the target domain and the subnetwork is trained from the \textit{same} initial values as in the source domain. 
    \item \textsc{Masks-Random:} Under this initialization strategy, \textit{only} the masks are used from the subnetwork obtained in the source domain. The parameters are randomly initialized from the distributions outlined in \S\ref{sec:init} before training on the target domain.
\end{itemize}

The results are shown in Figure \ref{fig:transfer}. Both \textsc{Masks-Reset} and \textsc{Masks-Random} show signs of generalization in the early stages of sparsification. Most notably, subnetworks obtained in the CDs domain are extremely robust; both the \textsc{Masks-Reset} and \textsc{Masks-Random} results show stronger performance than \textsc{Ticket-Target}, even in sparsity levels over 99\%. This is relatively surprising as the \textsc{Full-Model} in \S\ref{exp:obtain} achieved the worst performance in the CDs domain. Further inspection of representations learned in this domain will be required to understand its strong ticket performance, which may or may not be a coincidence. 

We see a 3-5\% dropoff in performance (up to 90\% sparsity) from tickets identified from the Books and Electronics tasks after transferring. These results together imply that tickets are not completely immune to distributional shifts, although the degradation in test accuracy is not substantial until reaching high sparsity. Nevertheless, we notice the accuracies of \textsc{Masks-Reset} and \textsc{Masks-Random} stay relatively stable from 0-90\% sparsity; they only begin to steadily decline after this point. 

Finally, we compare the performance of \textsc{Masks-Reset} and \textsc{Masks-Random}. In the Books tasks, \textsc{Masks-Random} performs better overall in comparison to \textsc{Masks-Reset}. Its performance is slightly worse in the Electronics and CDs tasks, although it is relatively comparable to \textsc{Masks-Reset} up to 96\%. Similar to the results in \S\ref{exp:obtain}, we notice a \textit{phase transition} point where the initial values (e.g., \textsc{Masks-Reset}) play a much bigger role in maintaining stability and performance in the deeper stages of sparsification.

\section{Discussion}

In this section, we briefly recap our findings, highlighting key points observed through our ticket procuring and transfer experiments. For each section, we also touch on areas for future work.

\paragraph{Evidence of transferability of winning tickets in natural language processing.}
Our experiments show that ``winning tickets'' can indeed be identified in a sentiment task formulated from noisy, user-generated datasets. Moreover, the ``winning tickets'', up to extreme level of sparsity (e.g., $\> 90\%$), can be transferred across domains without much loss in accuracy. The fact that tickets can be obtained in noisy environments shows its prominence across multiple data sources. However, our work only considers a binary sentiment analysis task. Future work can explore other tasks such as multi-class text classification, language modeling, and machine translation.

\paragraph{Randomly initialized tickets are strong baselines.} Consistent with the observations in \citet{liu2018rethinking}, initializing tickets to their \textit{original values} before training is not necessarily required for strong performance. In our experiments, we show that in high sparsity conditions (up to 90\%), there is no noticeable difference between the performance of the \textit{originally} and \textit{randomly} initialized subnetworks. Although the sparse masks build on top of each other from round $r_i$ to $r_{i+1}$, randomly initialized subnetworks are still able to settle in a local minima with comparable performance to that of the originally initialized subnetworks. However, our work fixes the optimizer and learning rate across experiments. It may be possible that randomly initialized subnetworks using varying optimization reach better minima. 

\paragraph{A \textit{phase transition} point largely influences ticket performance.} As alluded to above, there is almost no difference in performance when considering originally and randomly initialized subnetworks. However, our experiments point towards a crucial turning point---the \textit{phase transition}---in which the initialization begins to matter. In particular, especially in extreme levels of sparsity (e.g., 99.99\%) originally initialized networks exhibit less variance than randomly initialized tickets in test accuracy. However, the specific sparsity at which the phase transition happens is dataset-dependent. Understanding why this occurs and its relation with other models, datasets, and optimization algorithms can further unveil and explain the phenomena behind lottery tickets.  

\section{Applications in Federated Learning}

Federated learning is a scenario where a centralized model is trained over decentralized data, distributed across millions (if not billions) of clients (e.g., electronic devices) \cite{jakub2016federated,bonawitz2019towards}. Crucially, the clients are not allowed to exchange \textit{data} with the central server or each other. Instead, each client can fine-tune a model for a couple of iterations on their own data, then send their (encrypted) parameters or gradients to a server for aggregation. This ``collaborative learning" setup effectively maintains a level of user privacy by ensuring the data always stays on-device. However, this poses several challenges for optimization; as the centralized server does not have access to the data distribution of each client, any neural architecture selection has to be done on either (a) a \textit{different} data source the server has access to or (b) on each individual client. Since (b) is generally quite expensive, the server usually maintains some seed data, as alluded to in (a).

With the transferability of lottery tickets, the server can procure lottery tickets on server-accessible data, then retrain the tickets on client data under the federated learning framework. While there may be a large performance drop when transferring \textit{extremely} sparse networks, our results show that clients can still re-train \textit{moderately} sparse networks with commensurate performance. We believe that this ``sparsify and transfer" procedure has two immediate benefits: (1) past work---including the original incarnation of the lottery ticket hypothesis---has shown that sparse networks can be, under certain conditions, easier to optimize \cite{frankle2018the, morcos2019one, gale2019state}; and (2) sparser sub-networks have significantly less capacity than their large, over-parameterized counterparts, which can alleviate client-server communication costs (e.g., model uploading and downloading) \cite{jakub2016federated,sattler2019robust}.

\section{Conclusion}

The Lottery Ticket Hypothesis \cite{frankle2018the} posits that large, over-parameterized networks contain small, sparse subnetworks that can be re-trained in isolation with commensurate test performance. In this paper, we examine whether these tickets are robust against distributional shifts. In particular, we set up domain transfer tasks with the Amazon Reviews dataset \cite{mcauley2013hidden} to obtain tickets in a \textit{source} domain and transfer them in a disparate \textit{target} domain. Moreover, we experiment with the \textit{transfer} initialization of the networks, determining if resetting to initial values (obtained in the source domain) are required for strong performance in the target domain. Our experiments show that tickets (under several initialization strategies) can be transferred across different text domains without much loss up to a very high level of sparsity. 

In addition, there is a lot of debate on whether initial value resetting is critical to achieve commensurate test performance. While \citet{frankle2018the, frankle2019stable} present evidence supporting the importance of resetting, \citet{gale2019state, liu2018rethinking} show that sparse re-trainable subnetworks can be found independent of resetting. Our experiments show that this is \textit{not} a yes or no question. Specifically, we show there is a \textit{phase transition} related to sparsity. Resetting is not critical before extreme levels of sparsity (i.e., below 99\%), but the effect of resetting is magnified in high sparsity regimes. Finally, we demonstrate the practical applications of our results in federated learning.

\section*{Acknowledgments}

Thanks to Veselin Stoyanov and our anonymous reviewers for their helpful comments.

\bibliographystyle{acl_natbib_nourl}
\bibliography{emnlp2018}
\end{document}


\maketitle
\begin{abstract}
The Lottery Ticket Hypothesis \cite{frankle2018the} suggests that large and over-parametrized neural networks consist of small and sparse sub-networks that can be trained in isolation on the same dataset to reach a similar test accuracy with proper initialization. State-of-art neural network models for NLP tasks are often over-parametrized. This paper focuses on evaluating the performance of sparse sub-networks found in over-parametrized models under domain shift. Specifically, can a sparse sub-network obtained from a source dataset be transferred and re-trained in isolation on a target dataset in a different domain to match the performance of the sparse sub-network discovered on the target dataset? 
\end{abstract}

\section{Experiments}
\subsection{Hyperparameter Settings}
Our CNN uses three filters ($h \in [3,4,5]$), each with $127$ channels, and ReLU activation \cite{nair2010rectified}. We fix the maximum sequence length to $500$ words. The embeddings are $417$-dimensional and trained alongside the model. We opt not to use pre-trained embeddings to ensure the generalizability of our results. Additionally, we regularize the embeddings with dropout \cite{srivastava2014dropout}, $p=0.285$. The MLP contains one hidden layer with a dimension of $117$. Hyperparameters were discovered using Bayesian hyperparameter optimization \cite{snoek2012practical} on the Books validation set.

The model is trained with a batch size of $32$ for a maximum of $15$ epochs. Early stopping is used to save iterative model versions that perform well on a development set. We use the Adam optimizer \cite{kingma2014adam} with a learning rate of $1e^{-3}$ and $\ell_2$ regularization with a weight of $1e^{-5}$.

\bibliographystyle{acl_natbib_nourl}
\bibliography{emnlp2018}